\documentclass[letterpaper]{article} 
\usepackage{aaai25}  
\usepackage{times}  
\usepackage{helvet}  
\usepackage{courier}  
\usepackage[hyphens]{url}  
\usepackage{graphicx} 
\usepackage{natbib}  
\usepackage{caption} 
\usepackage{algorithm}
\usepackage{algorithmic}

\usepackage{amsmath}
\usepackage{amssymb}
\usepackage{mathtools}
\usepackage{amsthm}
\usepackage{booktabs}

\usepackage{float}

\usepackage{newfloat}
\usepackage{listings}
\DeclareCaptionStyle{ruled}{labelfont=normalfont,labelsep=colon,strut=off} 
\floatstyle{ruled}
\newfloat{listing}{tb}{lst}{}
\floatname{listing}{Listing}
\newcommand{\xhdr}[1]{{\noindent\bfseries #1}.}

\title{An Item is Worth a Prompt:  Versatile Image Editing with Disentangled Control}
\author{
    Aosong Feng\textsuperscript{\rm 1},
    Weikang Qiu\textsuperscript{\rm 1},
    Jinbin Bai\textsuperscript{\rm 2},
    Xiao Zhang\textsuperscript{\rm 3}\footnote{Corresponding author},
    Zhen Dong\textsuperscript{\rm 3},
    Kaicheng Zhou\textsuperscript{\rm 3},
    Rex Ying\textsuperscript{\rm 1}\footnotemark[1],
    Leandros Tassiulas\textsuperscript{\rm 1}\footnotemark[1]
}
\affiliations{
    \textsuperscript{\rm 1 }Yale University,
    \textsuperscript{\rm 2}National University of Singapore,
    \textsuperscript{\rm 3}Collov Labs
}

\begin{document}

\maketitle

\begin{abstract}
Building on the success of text-to-image diffusion models (DPMs), image editing has emerged as a crucial application for enabling human interaction with AI-generated content. Among various editing techniques, prompt-based editing has garnered significant attention for its capacity to simplify semantic control.
However, because diffusion models are typically pretrained on descriptive text captions, directly modifying words in text prompts often results in entirely different generated images, which undermines the objectives of image editing. 
Conversely, existing editing methods often employ spatial masks to maintain the integrity of unedited regions, but these are frequently disregarded by DPMs, leading to disharmonious editing outcomes.
To address these two challenges, we propose a method that disentangles the comprehensive image-prompt interaction into multiple item-prompt interactions, with each item associated with a uniquely learned prompt. 
The resulting framework, named \textbf{D-Edit}, leverages pretrained diffusion models with disentangled cross-attention layers and employs a two-step optimization process to establish item-prompt associations. 
This approach allows for versatile image editing by enabling targeted manipulations of specific items through their corresponding prompts.
We demonstrate state-of-the-art results in four types of editing operations including image-based, text-based, mask-based editing, and item removal, covering most types of editing applications, all within a single unified framework.
Notably, D-Edit is the first framework that can (1) achieve item editing through mask editing and (2) combine image and text-based editing.
We demonstrate the quality and versatility of the editing results for a diverse collection of images through both qualitative and quantitative evaluations.
\end{abstract}

%

\section{Introduction}
\label{sec:intro}

The recent advancements in text-to-image diffusion generative models represent a cutting-edge approach in the field of generative models. By gradually introducing noise into the image, these models facilitate sophisticated image synthesis \cite{podell2023sdxl,ruiz2023dreambooth,song2020denoising} while preserving semantic alignment with the text prompt.
One notable application is image editing, where diffusion models provide unprecedented control over various editing tasks, including inpainting \cite{nichol2021glide, avrahami2023blended}, text-guided editing \cite{hertz2022prompt,parmar2023zero}, pixel editing \cite{mou2023dragondiffusion, brooks2023instructpix2pix}, etc.
Various types of editing can generally be evaluated based on two key criteria: preservation of the original image's information and fidelity or consistency with the target guidance. An effective image editing process should prioritize retaining essential information from the original image while ensuring precise semantic alignment with the intended modifications.


\begin{figure*}[bpt]
  \centering \includegraphics[width=0.9\linewidth]{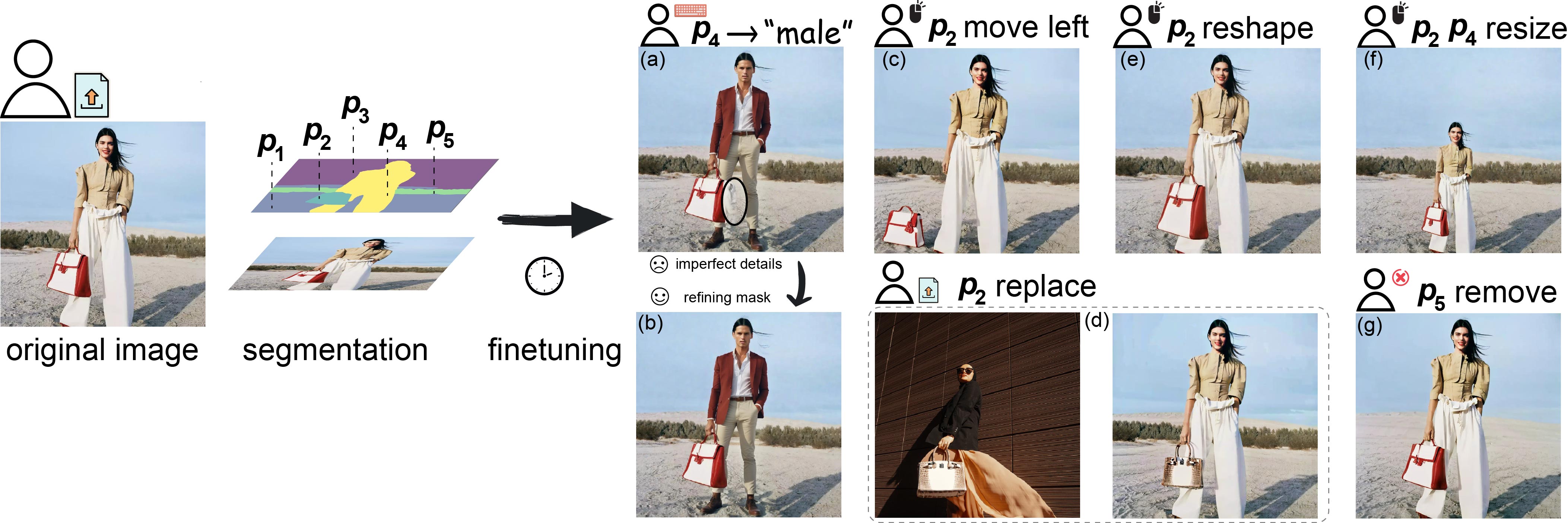}
  \caption{The editing pipeline of using D-Edit. 
  The user first uploads an image which is segmented into several items.
  After finetuning DPMs, the user can do various types of control, including
  (a) replacing the model with another using a text prompt;
  (b) refining imperfect details caused by segmentation;
  (c) moving bags to the ground;
  (d) replacing the handbag with another one from a reference image;
  (e) reshaping handbag;
  (f) resizing the model and handbag;
  (g) removing background.
  }
  \vspace{-10pt}
  \label{fig::pipeline}
\end{figure*}

To improve consistency with the target guidance, some work \cite{yang2023paint,chen2023anydoor, shen2023advancing,xue2022dccf} encodes reference images by introducing additional trainable encoders to preserve identities of the reference, and adds additional controls to DPMs using methods like ControlNet \cite{zhang2023adding}.
However, such methods cannot incorporate the existing text prompt control flow in DPMs and therefore require large-scale pretraining which is usually costly and domain-specific.
To preserve information about the original image and improve harmonization,
another line of work fixes diffusion sampling trajectory (by setting random seed or using DDIM) and achieves editing by carefully tuning text prompts \cite{mokady2023null,miyake2023negative}, changing part of the trajectory \cite{meng2021sdedit}, merging trajectories \cite{lu2023tf, wallace2023edict}, or optimizing the latent pixel space \cite{mou2023dragondiffusion, shi2023dragdiffusion}.
This avoids additional pretraining but either relies on careful source prompt design to match the editing region or additional optimization per edit.


In this work, we propose two key techniques aimed at enhancing the aforementioned criteria:
(1) \textbf{Disentangled Control}: To preserve the original image's information, the editing of a target item should minimally impact surrounding items. The control process from prompt to image should also be disentangled, ensuring that modifications to an item's prompt do not interfere with the control flow of other items. Recognizing that text-to-image interactions occur within the cross-attention layers of attention-based diffusion models, we propose a grouped cross-attention mechanism to disentangle the control flow between prompts and items.
(2) \textbf{Unique Item Prompt}: 
To enhance consistency with the guidance (e.g., a reference image), each item should be associated with a unique prompt that directs its generation. These prompts typically involve special tokens or rare words. Previous works on image personalization, such as Dreambooth \cite{ruiz2023dreambooth} and Textual Inversion \cite{gal2022image}  have explored this concept by representing a new subject with a unique prompt, which is then used for image generation.
In contrast, our approach employs independent prompts to define individual items rather than the entire image. Ideally, if each item in the image, with all its details, could be precisely described by a unique English word, users could achieve various editing tasks simply by swapping the current word for the desired one.


By fully harnessing the potential of prompt uniqueness and disentangled control, we introduce a versatile image editing framework for diffusion models called Disentangled-Edit (D-Edit). This unified framework enables a wide range of image editing operations at the item level, including text-based, image-based, mask-based editing, and item removal. As illustrated in Fig. \ref{fig::pipeline}, the process begins with segmenting the target image into multiple editable items (in this context, background and unsegmented regions are also referred to as items), each associated with a prompt composed of several new tokens. 
The associations between prompts and items are established through a two-step fine-tuning process, which optimizes both the text encoder's embedding matrix and the UNet model's weights.
To disentangle prompt-to-item interactions, we introduce grouped cross-attention, which isolates attention calculation and value updates. This allows users to achieve various types of image editing by modifying prompts, items, and their associations, as well as by adjusting corresponding masks. 
This flexibility opens up a wide range of creative possibilities and offers precise control over the editing process. We demonstrate the versatility and performance of our framework across four image editing tasks, utilizing both Stable Diffusion and Stable Diffusion XL.
We summarize our contribution as follows:
\begin{itemize}
\item We propose to establish item-prompt association to achieve item editing.
\item We introduce grouped cross-attention to disentangle the controlling flow in diffusion models.
\item We propose D-Edit as a versatile framework to support various image-editing operations at the item level, including text-based, image-based, mask-based, editing, and item removal.
D-Edit is the first framework that can do mask-based editing, and perform text and image-based editing at the same time. Code can be found at https://github.com/collovlabs/d-edit
\end{itemize}

\begin{figure*}[ptb]
  \centering \includegraphics[width=0.85\linewidth]{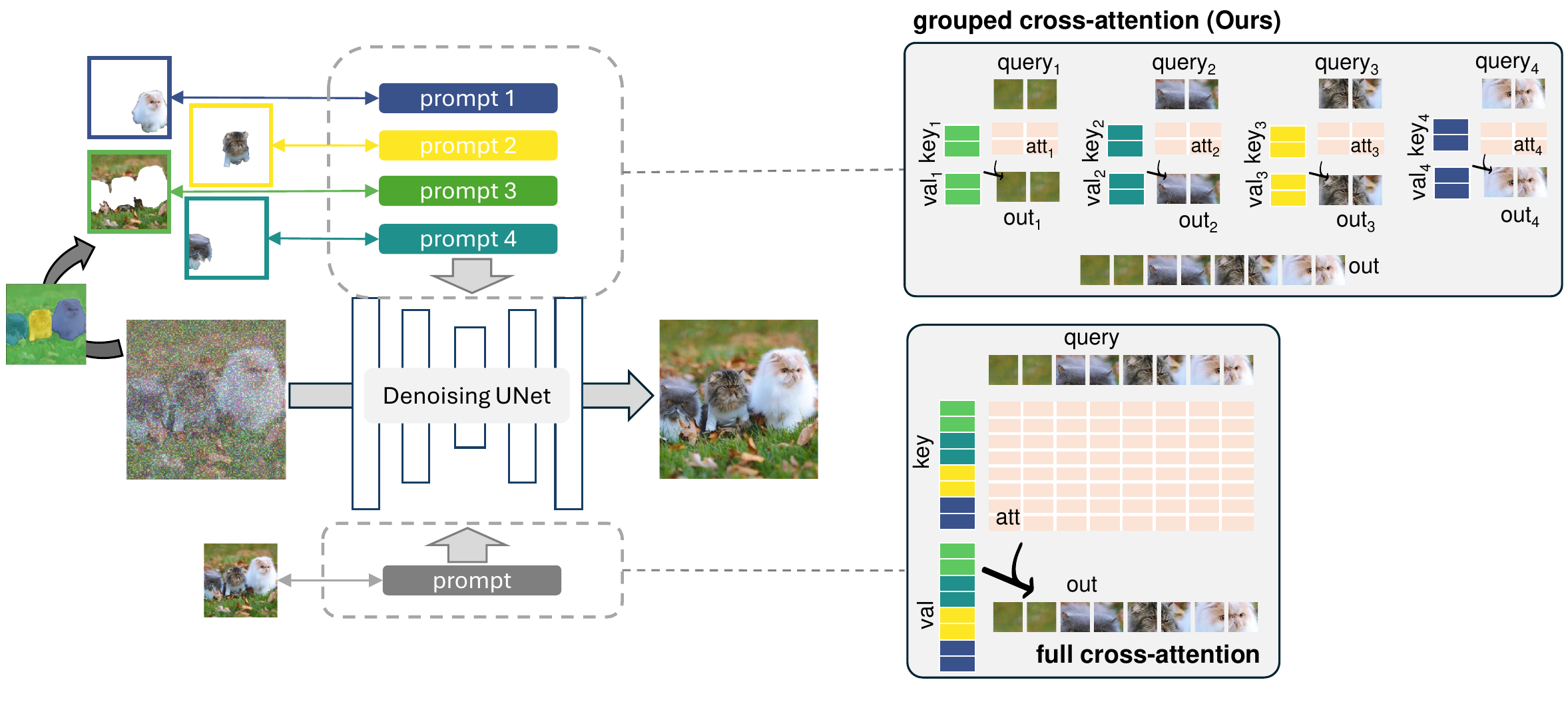}
  \vspace{-5pt}
  \caption{Comparison of conventional full cross-attention and grouped cross-attention.
  Query, key, and value are shown as one-dimensional vectors.
  For grouped cross-attention, each item (corresponding to certain pixels/patches) only attends to the text prompt (two tokens) assigned to it.
  }
  \vspace{-7pt}
  \label{fig::promptitem}
\end{figure*}

\section{Related Works}
\label{sec:related}

\xhdr{Trajectory-Based Editing}
Because natural language cannot perfectly describe a given image, a single prompt may correspond to multiple sampling trajectories with different random seeds.
SDEdit \cite{meng2021sdedit} achieves editing by sharing the former part of the sampling process to preserve the high-level information like layout and changing the latter part for realistic reconstruction.
Diffusion inversion \cite{mokady2023null, huberman2023edit, lu2023tf} inverts the reverse diffusion process and forces the original and edited trajectory to share the same sampling starting point (picky to the sampling method).
Interactions between the two trajectories can then be built by sharing cross/self-attention to preserve the original identity \cite{tumanyan2023plug, mou2023dragondiffusion}.
Combined with P2P \cite{hertz2022prompt}, these methods can achieve fine-grained text-based editing, but it requires an accurate captioning prompt of the original image to reverse the diffusion, and the prompt can only change by a few words.
Our methods are agnostic to sampling trajectories, don't require any prior prompt, and support more freedom to change the prompt.

\xhdr{Image Identity Extraction}
Because image-based editing involves additional modalities for conditioning, the natural thought is to introduce additional encoders to encode the corresponding modalities.
Paint-by-example \cite{yang2023paint} trains additional MLP layers following a pretrained CLIP image encoder to encode reference image information. 
AnyDoor \cite{chen2023anydoor} design an additional identity extractor to preserve the original item identity.
PCDMs \cite{shen2023advancing} introduced additional layers to encode the source image and target position.
Because of the introduction of the additional trainable module, these models have to train on a large dataset to perform well on more images. 
Our method leverages the original pretrained text-encoder and UNet to encode reference images and can therefore process a wider range of images.

\xhdr{Image as a Word}
Representing images with special tokens has been a popular choice for image personalization.
Textual Inversion \cite{gal2022image} and Dreambooth \cite{ruiz2023dreambooth} represent the original subject with new tokens or rare tokens, the embedding layers or the full model are optimized with a few personalization optimization steps.
We follow this line of thought in the image editing context. Instead of learning prompts from images, we learn from items and therefore can be applied when the given image contains multiple items with different subjects.
The most similar works to us are SINE \cite{zhang2023sine} and Imagic \cite{kawar2023imagic}.
SINE achieves single-image editing by combining Dreambooth-trained prompt with source prompt using classifier-free guidance.
Imagic optimizes the prompt embedding to be aligned with both the input image and the target text and interpolates the learned prompt to achieve editing.
Compared to these methods, our framework does not require a captioning prompt in advance and can achieve more types of controls besides text-based editing.

\section{Method}
In this section, we discuss the details of the D-Edit framework.
We first review the basics of diffusion models and text-to-image control flow.
Next, we show how to establish item-prompt association through the two-step finetuning.
Then, we discuss how to utilize the editability of prompts for versatile image editing operations.

\subsection{Diffusion Models}
 Denoising diffusion probabilistic models generate high-quality images by learning to reverse the given forward Markov chain through iterative refinement.
During the forward process, it works by gradually adding Gaussian noise to the original data, deriving intermediate latent as
\begin{equation}
\vspace{-1.2 pt}
\begin{split}
z_t = \sqrt{\alpha_t}x_0 + \sqrt{1-\alpha_t} \epsilon_t
\end{split}
\end{equation}
with $0=\alpha_T<\alpha_{T-1}<..<\alpha_0=1$ being the noise schedule, and $\epsilon_t \sim \mathcal{N}(0,\mathbb{I})$.
The neural network $f_\theta(z_t, t)$ (like UNet) is introduced to predict the added noise $\epsilon_t$.
The predicted noise is then used for sampling by running the reverse process which starts from the pure Gaussian noise $z_T$ and ends at original data $z_0$.
Latent Diffusion Model (LDM) is the most widely adopted diffusion model for high-resolution image generation. 
Given an image $I\in\mathbb{R}^{H\times W \times 3}$, LDM operates in the encoded latent space, $z_0 = E(I)$, and maps the sampled latent representation to the original space using the paired decoder.

\xhdr{Text-to-Image Control} A key factor contributing to the success of LDM is its robust ability for text-to-image generation.
By introducing the additional condition $y$ as the auxiliary input to $f_\theta(z_t,t,y)$, LDM can generate images according to the designed user prompt. 
It should be noted that such prompts are usually general textual descriptions, and the final generated image is additionally controlled by the random seed in use for sampling. 

Textual prompt controls the image generation through the cross-attention process.
Specifically, the given text prompt $P$ containing $W$ words is first encoded by the pretrained text encoder (e.g. CLIP \cite{radford2021learning}) $g_\phi$ into text embedding $c=g_\phi(P) \in \mathbb{R}^{W\times D_c}$ ($W$ is the prompt length and $D_c$ is the embedding dimension). 
It is then used as input along with the image latent $z_t \in \mathbb{R}^{Z\times D_z}$ (we abuse the notation of model input and layer input) in the UNet cross-attention layer:
\begin{equation}
\label{eq:att}
\begin{split}
q&= w_q z_t \in \mathbb{R}^{Z\times D} \\
k&= w_k c  \in \mathbb{R}^{W\times D}\\
v&= w_v c \in \mathbb{R}^{W\times D} \\
\end{split}
\qquad
\begin{split}
  &A = \text{softmax}(qk^T) \in \mathbb{R}^{Z\times W} \\
 &O(c, z_t) = A\cdot v,
\end{split}
\end{equation}

where the condition $c$ is encoded into key and value vector while the image input $z_t$ is encoded into query vector.

\subsection{Item-Prompt Association}

As shown in Eq. \ref{eq:att}, the original LDM performs text-image interaction between every token in $c$ and every pixel in $z_t$ through cross-attention matrix $A$.
In fact, such token-pixel interactions have been shown disentangled in nature \cite{tang2022daam, hertz2022prompt}, and the attention matrix $A\in\mathbb{R}^{Z\times W}$ is usually sparse in the sense that each column (token) only attend to several non-zero rows (pixels).
For example, during image generation, the word "bear" has higher attention scores with pixels related to the bear region compared to the remaining region.

Inspired by the natural disentanglement, we propose to segment the given image $I$ into $N$ non-overlapped items $\{I_i \}_{i=1}^{N}$ using segmentation model (same segmentation applied to $z^t$ because of emergent correspondence \cite{tang2023emergent}).
A set of prompts $\{ P_i\}_{i=1}^{N}$ is adopted to replace the original text prompt $P$.
Each prompt $P_i$ is regarded as the textual representation of the corresponding item $z^t_i$ (Details of $P_i$ are discussed in Sec. \ref{sec::item-prompt}).
As shown in Fig. \ref{fig::promptitem},
we force different items $I_i$ to be controlled by distinct prompt $P_i$ by masking our other items, and therefore any prompt changes in $P_i$ will not influence the remaining item during the cross-attention controlling flow, which is the desired property for image editing.
This results in a group of disentangled cross-attentions. For each item-prompt pair  ($I_i$, $P_i$), the cross-attention can be written as
\begin{equation}
\label{eq::dis_ca}
\begin{split}
    &q_i=w_q z^t_i \in \mathbb{R}^{Z_i\times D} \\ 
    &k_i = w_k c_i  \in \mathbb{R}^{W_i\times D} \\
    &v_i = w_v c_i \in \mathbb{R}^{W_i\times D}\\
\end{split}
\quad
\begin{split}
&\text{out}(\{ c_i\}, \{z^t_i\}) = \Sigma_{i=1}^{N} \text{out}_i(c_i, z^t_i)\\
    &A_i = \text{softmax}(q_ik_i^T) \in \mathbb{R}^{Z_i\times W_i} \\
    & \text{out}(c_i, z^t_i) = A_i\cdot v_i
\end{split}
\end{equation}
It should be noted that such disentangled cross-attention cannot be directly used for pretrained LDMs, and therefore further finetuning is necessary to enable the model to comprehend item prompts and grouped cross-attention. 

\subsection{Linking Prompt to Item}
\label{sec::item-prompt}
We link prompts to items with two sequential steps. 
We first introduce the item prompt, consisting of several special tokens with randomly initialized embeddings.
Then we finetune the model to build the item-prompt association.


\begin{figure}[tbp]
  \centering
 \includegraphics[width=0.8\linewidth]{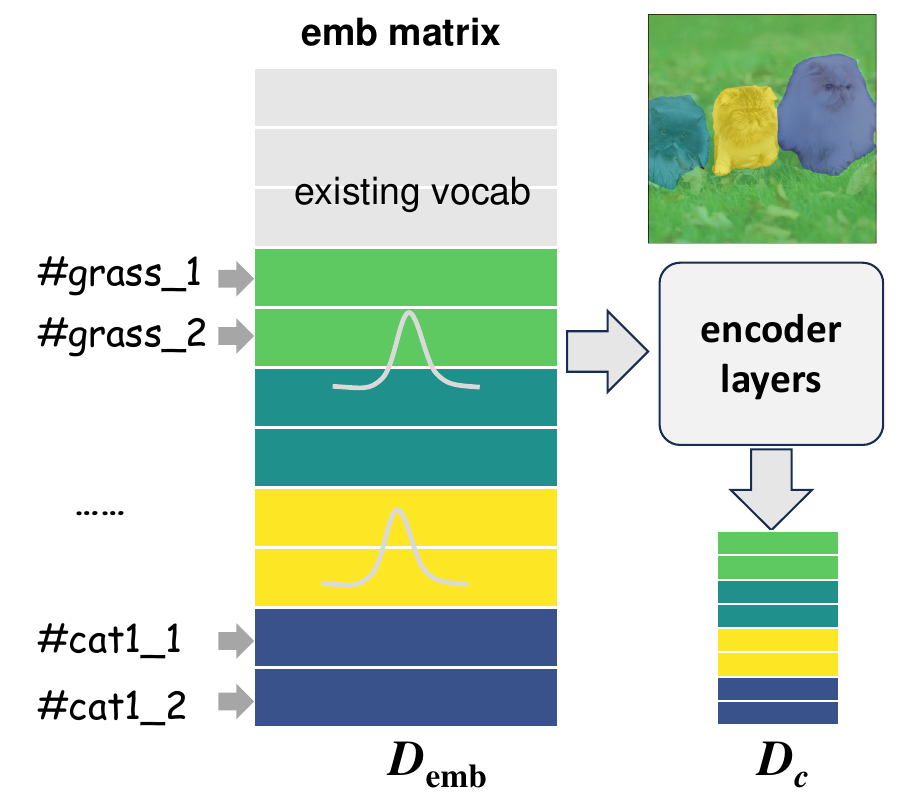}
 \vspace{-5pt}
  \caption{Embedding layer in the text encoder. New tokens are inserted with random initialization. 
  }
   \vspace{-8pt}
  \label{fig::textencoder}
\end{figure}

\xhdr{Prompt Injection} 
We propose to represent each item in an image with several new tokens which are inserted into the existing vocabulary of text encoder(s).
Specifically, as shown in Fig. \ref{fig::textencoder}, we use 2 tokens to represent each item and initialize the newly added embedding entries using Gaussian distribution with mean and standard deviation derived from the existing vocabulary.
For comparisons, Dreambooth \cite{ruiz2023dreambooth} represents the image using rare tokens and
perfect rare tokens should have no interference with existing vocabulary, which is hard to find.
Textual inversion and Imagic insert new tokens into vocabulary where the corresponding embedding is semantically initialized by given word embeddings which describe the image. This adds additional burdens of captioning the original image. 
We found that it is sufficient to use randomly initialized new tokens as item prompts and such randomly initialized tokens have minimal impact on the existing vocabularies.

To associate items with prompts, the inserted embedding entries are then optimized to reconstruct the corresponding image to be edited using
\begin{equation}
\label{eq::opt}
    \text{min}_e \mathbb{E}_{t,\epsilon}\left[|| \epsilon - f_\theta (z_t, t, g_\Phi(P) )||^2  \right],
\end{equation}    
where $e\in\mathbb{R}^{NM\times D_{\text{emb}}}$ 
represents the embedding rows corresponding to $N$ items each with $M$ tokens.


\xhdr{Model Finetuning }
Optimization in the first stage injects the image concept into text-encoder(s), but cannot achieve perfect reconstruction of the original item given the corresponding prompt. 
Therefore, in the second stage of optimization, we optimize the UNet parameters by running optimization with the same objective function as in Eq. \ref{eq::opt}.
We found that updating parameters solely within cross-attention layers is adequate, as we only disentangle the forward process of these layers rather than the entire model.
It should be noted that the optimizations above are running against only one image or two images (target and reference images) if image-based editing is needed.

\subsection{Editing with Item-Prompt Freestyle}
After the two-step optimization, the model can exactly reconstruct the original image given the set of prompts corresponding to each item, with an appropriate classifier-free guidance scale.
We then achieve various disentangled image editing by changing the prompt associated with an item, the mask of an item-prompt pair, and the mapping between items and prompts.
We discuss four types of image editing operations that can be achieved by varying item-prompt relationships, summarized in Fig. \ref{fig::editingmask}. Details of each operation are discussed in Appendix.

\begin{figure}[tbp]
  \centering \includegraphics[width=0.8\linewidth]{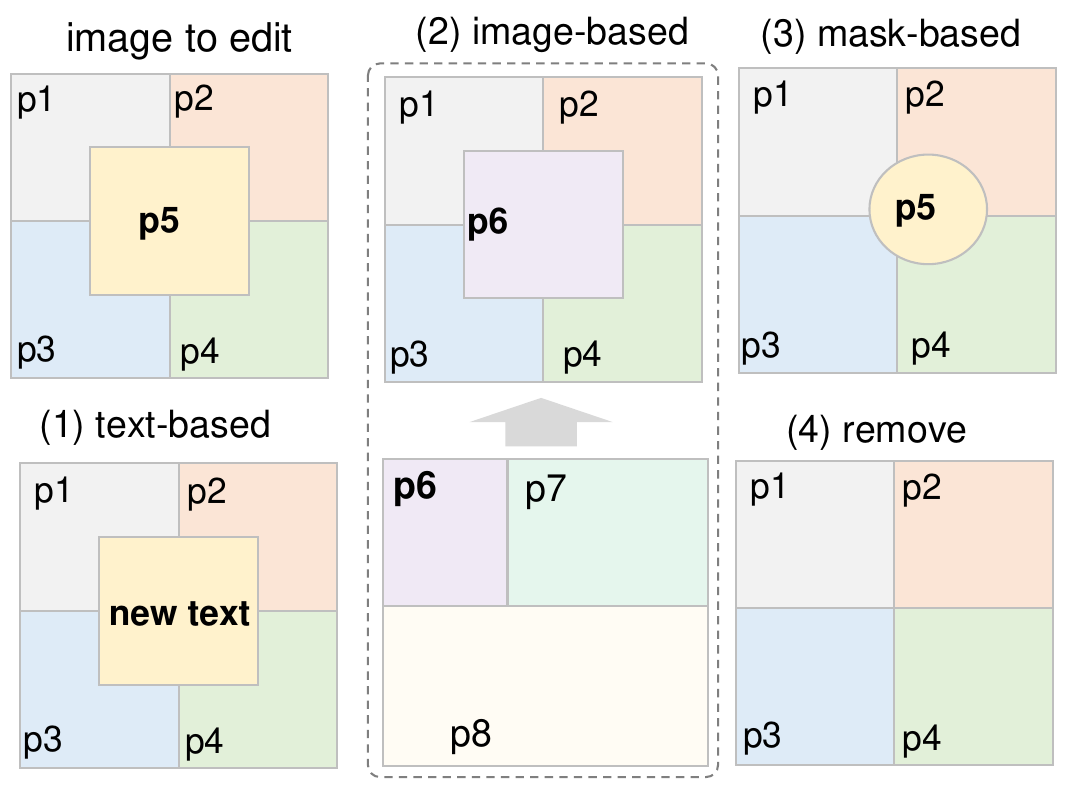}
  \caption{Operations needed for different types of image editing.
  Each colored item has a unique prompt p.}
  \label{fig::editingmask}
    \vspace{-9pt}
\end{figure}




\section{Experiments}
\subsection{Experiment Setup}

\xhdr{Training Details} We implement our D-Edit framework on stable diffusion (SD) v1.5 for 512 $\times$ 512 images and SDXL for 1024 $\times$ 1024 resolution images.
Mask2Former \cite{cheng2022masked} is used for segmentation and Grounding DINO \cite{liu2023grounding} for text-prompted segmentation.
The finetuning is performed with Adam optimizer with learning rate 1\textit{e}-4 for embedding training, 5\textit{e}$-$5 for cross-attention layer training, and 5\textit{e}$-$5 for LORA full parameter training.
Gradient accumulation is applied to keep the effective batch size to 10 for training robustness.
Each image is segmented into 3-8 items by merging excess segments and each item is represented by 1 token for SD and 5 tokens for SDXL.
We deploy the default Euler discrete scheduler with sampling step 20 to generate all images during inference.
All finetuning and inference are conducted on NVIDIA A6000 GPUs.

\begin{figure}[hptb]
  \centering \includegraphics[width=1\linewidth]{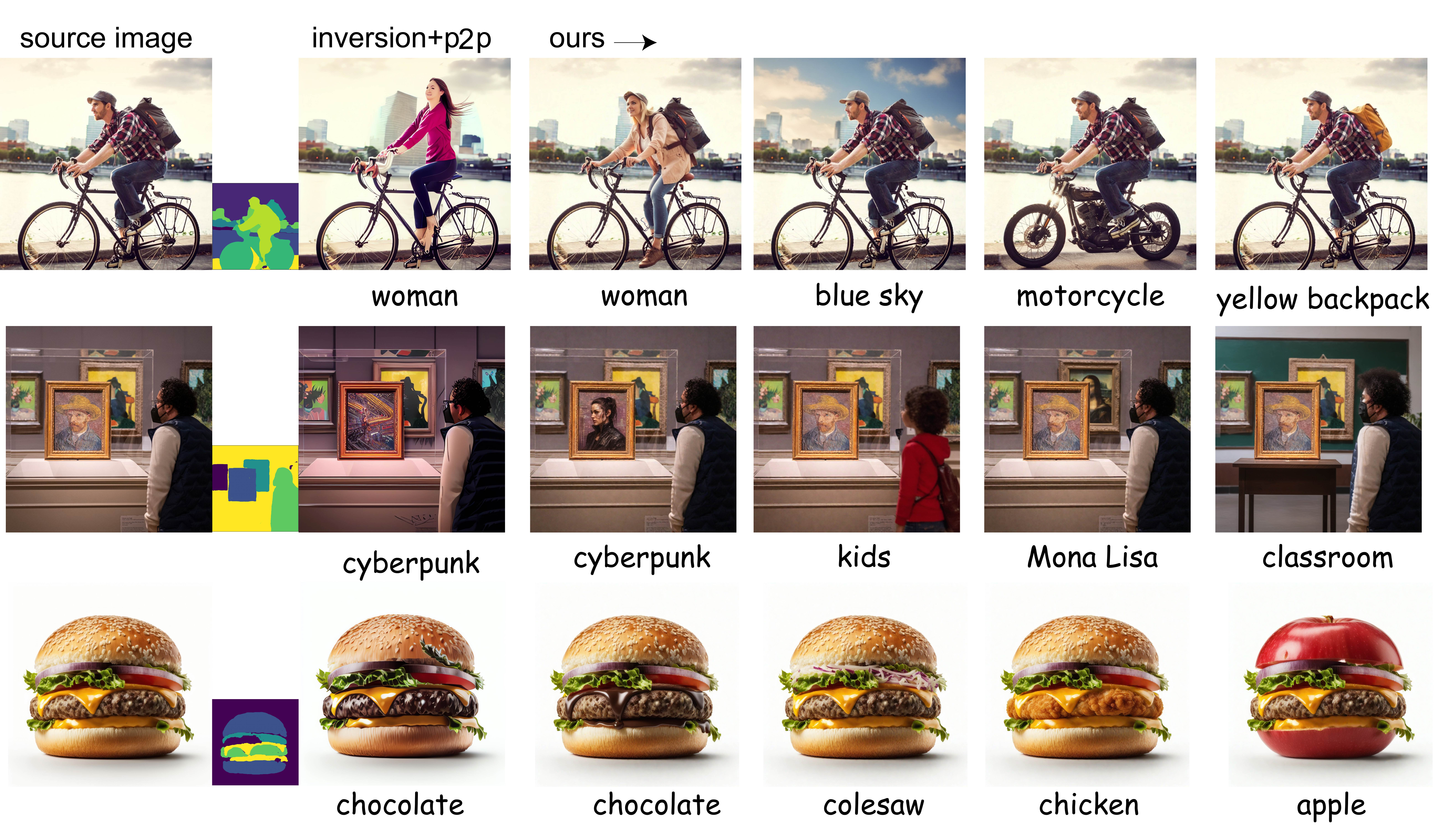}
  \caption{Text-guided editing. D-Edit enables selection of any item segmentation and edit using text prompt.}
  \vspace{-10pt}
  \label{fig::text1}
\end{figure}

\subsection{Text-Guided Editing}
Given an input image with appropriate segmentation (no captions needed),
we can select any one of the items and replace the learned prompt with the target text prompt.
We show such text-guided editing results in Fig. \ref{fig::text1}.
Compared to null-text inversion with Prompt-to-Prompt (P2P), D-Edit can generate more realistic details and have a more natural transition between edited and unedited areas (e.g., the connection between the bike handlebar and the woman's hand) because of disentangled control.
The editing of D-Edit is more focused on the target item while the editing of inversion with P2P overflows to other regions (painting in the second example and cheese in the third) using text.
Besides, unlike most text-guided methods, D-Edit does not require a caption for the original image which is extremely useful when the scene is hard to describe.

\begin{figure}[h]
  \centering \includegraphics[width=1.0\linewidth]{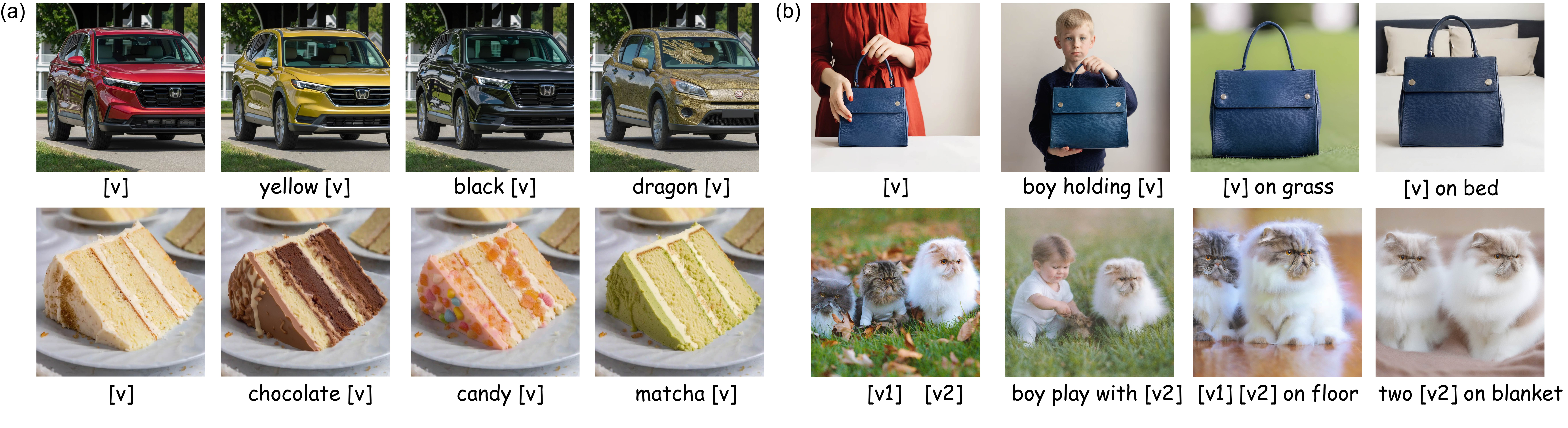}
  \caption{
   The learned prompt (denoted as [v]) can be combined with words to achieve refinement/editing of the target item.
  (a) Augment an item prompt with words while keeping other prompts unchanged for editing.
  (b) Generate the entire image with certain item prompt(s) augmented with text words for personalization.
  }
  \label{fig::text2}
\end{figure}

\begin{figure*}[hptb]
  \centering \includegraphics[width=0.98\linewidth]{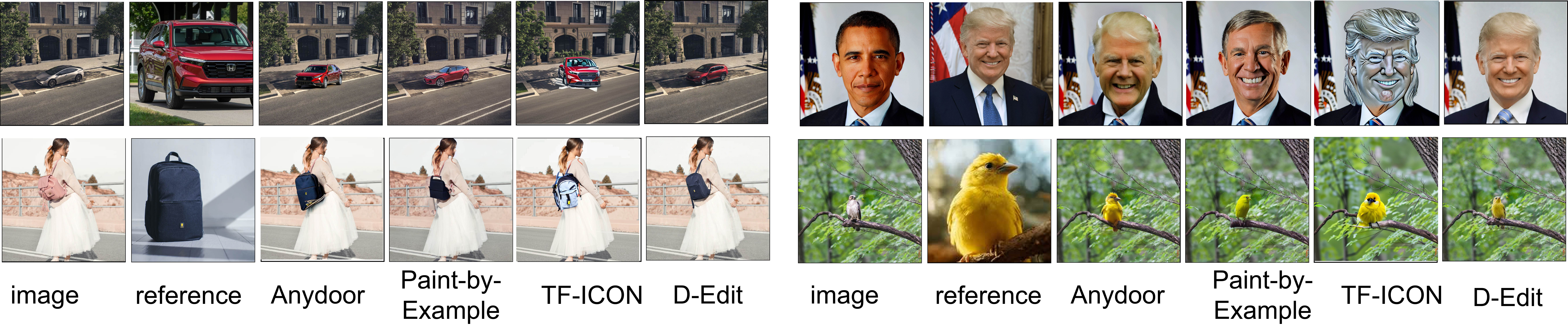}
  \caption{Qualitative comparison of image-guided editing. D-Edit is compared with Anydoor, Paint-by-Example, and TF-ICON, on item replacement and face swapping.
  }
  \vspace{-10pt}
  \label{fig::image2}
\end{figure*}

Then we show the learned item prompts can be combined with normal text words to achieve item refining, besides item replacement. 
As shown in Fig. \ref{fig::text2}(a), by combining the learned prompt with adjective words, we can achieve color and texture control of specific items.
The preservation of the car and cake shape details after editing indicates the established association between prompts and items through finetuning, while the add-on effect shows the good quality of the learned new prompt in the vocabulary.

In Fig. \ref{fig::text2}(b), we show the results of item personalization where we generate the image using certain item prompt(s), without using the item-prompt association.
This differs from Dreambooth-style personalization in that Dreambooth lacks the capability for item-level customization unless the item is cropped and focused upon, which is usually hard in an image including multiple items. Besides, it requires more images (3-5) with captions for personalization, while our method takes one image without captions.
Qualitative results show that learned item(s) can be combined with text to generate personalized images, and the text prompt can be used to personalize the background, number, and position of the item.

For quantitative evaluation, we introduce a new benchmark D-Item(Text) with 100 manually selected multi-item images, where each image is properly segmented into 3-8 items with a segmentation mask. We also include the caption of each image used for baselines, although it is not needed for D-Edit.
2 items of each image are selected and given 5 appropriate target prompts, therefore generating 1,000 item-prompt pair combinations.
We adopt CLIP text (CLIP-T) score to represent the semantic alignment of the edited item and target prompt, and LPIPS score to represent consistency with the original images.
As shown in Tab. \ref{tab::text}, D-Edit outperforms SDEdit \cite{meng2021sdedit} and P2P with DDIM inversion, especially on LPIPS score which shows improved fidelity to the original images.

\begin{table}[h]
\centering
\caption{Text-guided editing: consistency with the original image (LPIPS) and target text prompt (CLIP-T).}

\resizebox{0.45\columnwidth}{!}{
\begin{tabular}{lcc} 
\toprule
       & LPIPS$\downarrow$               & CLIP-T$\uparrow$                 \\ 
\midrule
P2P    & 0.401                           & 39.5                           \\
SDEdit & 0.432                           & 32.1                            \\
D-Edit & \textbf{0.179} & \textbf{42.0}  \\
\bottomrule
\end{tabular}
}

\vspace{-5pt}
\label{tab::text}
\end{table}

\begin{table}[h]
\centering
\caption{ Image-guided editing: consistency with the original image (LPIPS$_t$) and reference image (LPIPS$_t$ and CLIP-I).}
\resizebox{0.55\columnwidth}{!}{
\begin{tabular}{lccc} 
\toprule
                 & LPIPS$_t$$\downarrow$ & LPIPS$_r$$\downarrow$ & CLIP-I$\uparrow$  \\ 
\midrule
Aydoors          & 0.608                & 0.720                & 60.2              \\
PbE & 0.465                & 0.833                & 50.8              \\
D-Edit           & $\textbf{0.340}$     & $\textbf{0.701}$     & $\textbf{66.4}$   \\
\bottomrule
\end{tabular}
}
\label{tab::image_scores}
\end{table}

\subsection{Image-Guided Editing}

For image-guided editing, the user can select one item from the reference image and use it to replace one item in the target image.
We compare the editing results with baselines including Anydoors \cite{chen2023anydoor}, Paint-by-Example \cite{yang2023paint} and TF-ICON \cite{lu2023tf} when the reference image mainly consists of a single item.
As shown in Fig. \ref{fig::image2},
Paint-by-example can naturally inpaint reference items into the target scene but it falls short in keeping the identity of the reference item (the face and bird example).
Anydoors can retain more relevant details from the reference image, yet it may also incorporate undesirable elements in reference, resulting in a less harmonious blend with the target image.
For example, the car's original orientation is preserved, causing it to appear out of the parking spot in the target image.
Besides, it cannot preserve the face details as in the example.
Compared to these methods, 
D-Edit is capable of seamlessly composing objects into the target while maintaining their identities.

\begin{figure*}[pt]
  \centering \includegraphics[width=0.9\linewidth]{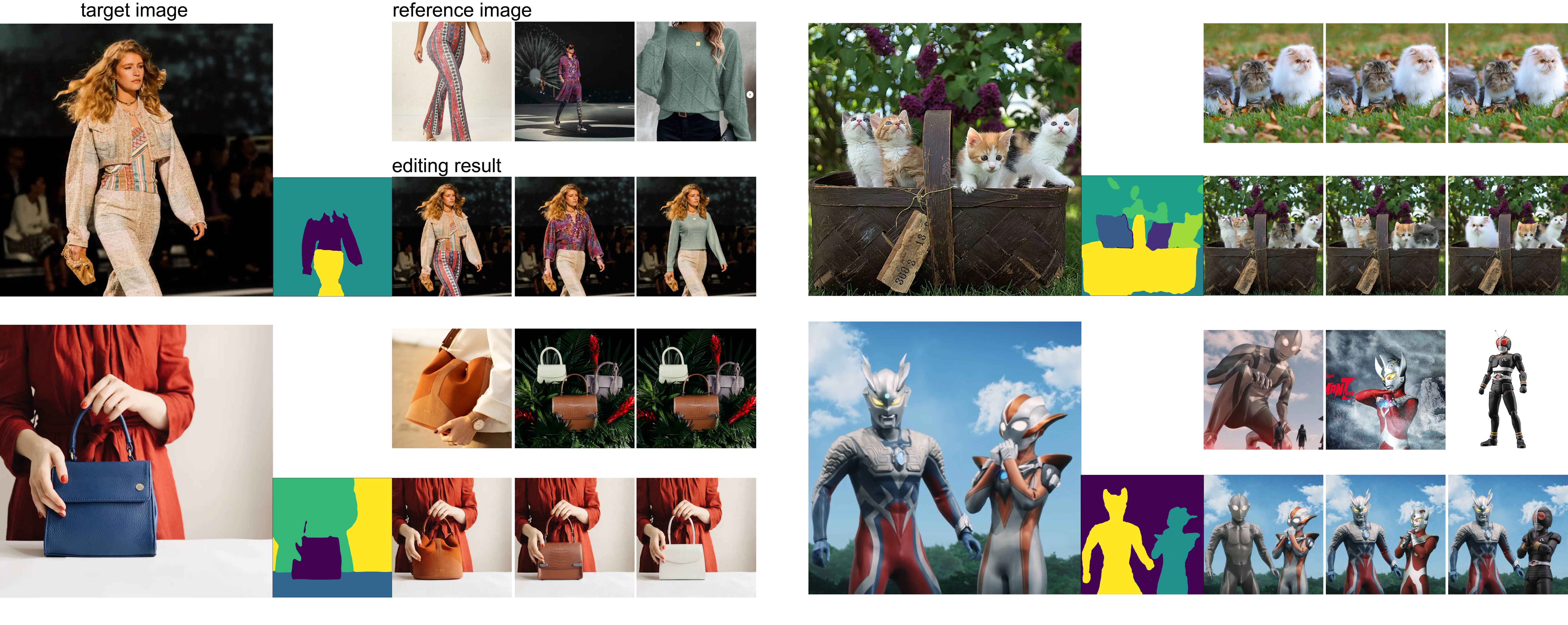}
  \vspace{-5pt}
  \caption{
    Image-guided editing:
    Any item in the image can be replaced by another item from the same or different images.
  }
  \vspace{-10pt}
  \label{fig::image1}
\end{figure*}

We show more image-based editing results of D-Edit in Fig. \ref{fig::image1} and Appendix. 
D-Edit can work well when the reference image contains multiple items that may be hard to separate (like the bag in hand).
Additionally, D-Edit doesn't necessitate the reference item to closely resemble its anticipated appearance in the target image, because blending through the prompt space offers smoother transitions compared to pixel-level manipulation, and the prompt-mask correspondence helps standardize the appearance of the reference item.
For example, in the Ultraman example, the reference and target Ultraman can take completely different postures (kneeling v.s. standing).

For quantitative evaluations, based on D-Item(Text) benchmark, we then construct the D-Item(Image) benchmark where each selected item is paired with two reference items from two different reference images, resulting in 400 item-item pairs.
Three metrics are considered: LPIPS$_t$ measures consistency with the original target image; LPIPS$_r$ and CLIP-Image (CLIP-I) measure alignment with the reference image in low- and high-level feature spaces.
As shown in Tab. \ref{tab::image_scores}, both D-Edit and Paint-by-Example can achieve high fidelity to the original image, while D-Edit can also preserve the target image better compared with Anydoors.

\subsection{Mask-Based Editing and Item Removal}
\begin{figure}[htb]
  \centering \includegraphics[width=1.0\linewidth]{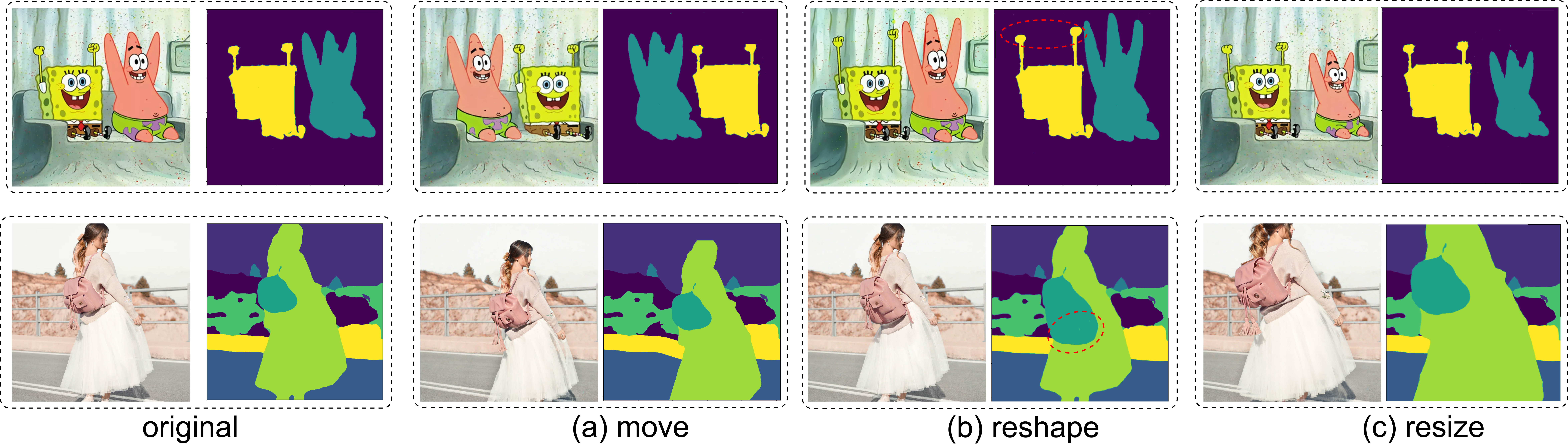}
    \vspace{-8pt}
  \caption{Different types of mask-based editing: (a) Moving/swapping items; (b) reshaping an item; (c) Resizing an item.}
   \vspace{-8pt}
  \label{fig::mask_guided}
\end{figure}

\begin{figure}[h]
  \centering \includegraphics[width=1.0\linewidth]{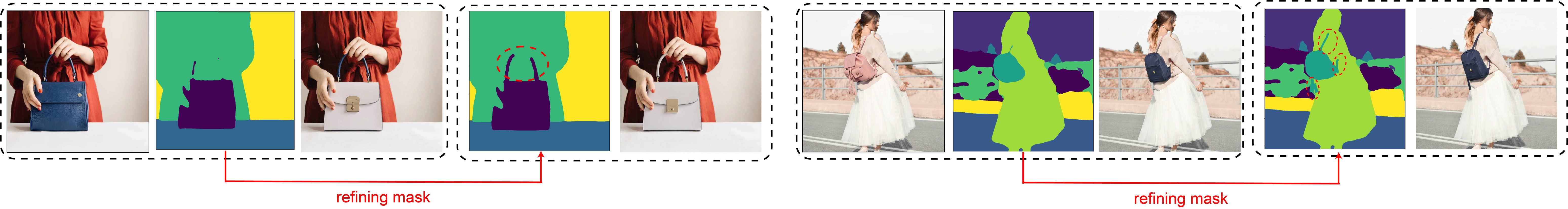}
    \vspace{-8pt}
  \caption{Post-editing refinement can be performed when obtaining imperfect results due to imperfect segmentation.}
   \vspace{-8pt}
  \label{fig::post_edit}
\end{figure}

For mask-based editing, we explore four types of operations on the target items, including moving, reshaping, resizing, and refinement.
As shown in Fig. \ref{fig::mask_guided}, D-Edit can edit the shape of the target item by simply editing the corresponding mask.
Because of the mask-item-prompt association, the disentangled attention can imagine and fill the new details in the edited regions according to the given item prompt, therefore leading to natural editing results.
We also show the post-editing performance in Fig. \ref{fig::post_edit}.
This can be useful when initial masks from the segmentation model do not cover the whole item, like the missing handle of the handbag and missing straps of the backpack, which will lead to imperfect (image/text/mask-guided) editing results.
D-Edit can later fix these mask details and regenerate using the same random seed, and lead to refined results.

\begin{figure}[hb]
  \vspace{-8pt}
  \centering \includegraphics[width=1.0\linewidth]{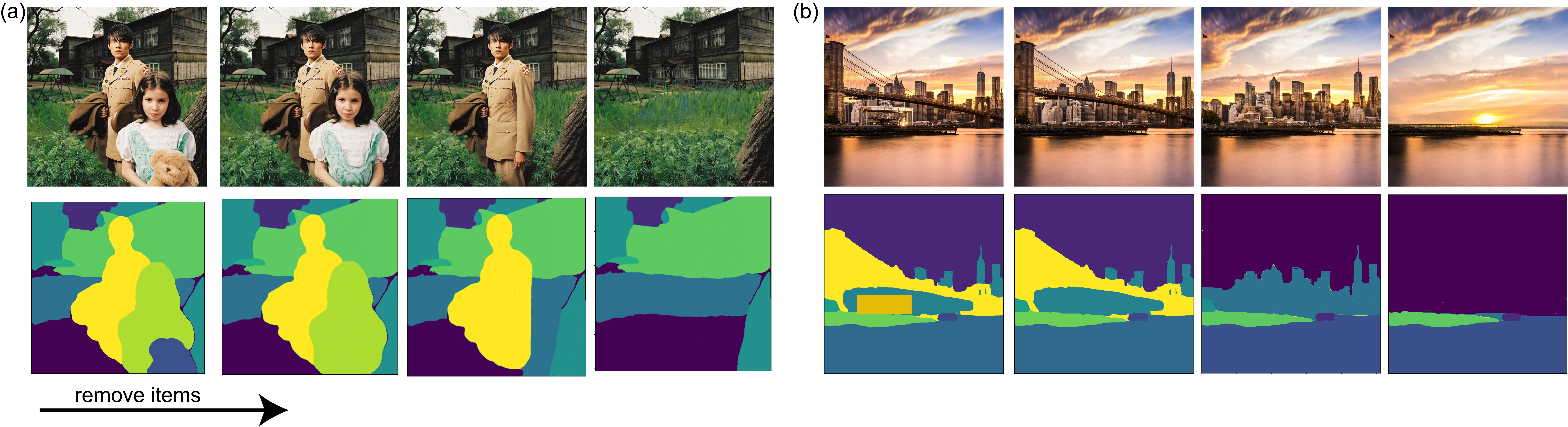}
  \caption{ Removing items one by one from the image.}
  \label{fig::remove}
\end{figure}

D-Edit also enables removing items by deleting the mask-item-prompt pairs.
In Fig. \ref{fig::remove}, by deleting items from the scene image one by one, the resulting blank region will be re-partitioned to nearby masks and join the corresponding item-prompt pair.
D-Edit will then use such new associations to imagine the blank regions and therefore lead to reasonable filling results.
More visual results can be found in Appendix.
To quantitatively assess the item-removed images, we conducted user studies with a group of 15 annotators. 
They are asked to score 30 pairs of original and item-removed images from 1 to 5 (higher means better), based on quality (how well the region after removal harmonized with the surrounding scene) and fidelity (the reasonableness of the filling content).
D-Edit is compared with SDXL inpaint model by inpainting the region where the item is removed with the surrounding item's caption, and results are shown in Tab. \ref{tab::user_remove}.

\begin{table}[hb]
\centering
\caption{Quality and fidelity of editing after removing items.}
\resizebox{0.5\columnwidth}{!}{
\begin{tabular}{lccc} 
\toprule
                 & Quality $\uparrow$ & Fidelity $\uparrow$  \\ 
\midrule
SDXL-inpaint &  3.26      & 2.42           \\
     D-Edit  & \textbf{4.01}     & \textbf{4.44}      \\
\bottomrule
\end{tabular}
}
\label{tab::user_remove}
\end{table}

\subsection{Ablation Study}
\begin{figure}[hb]
  \vspace{-8pt}
  \centering \includegraphics[width=1.0\linewidth]{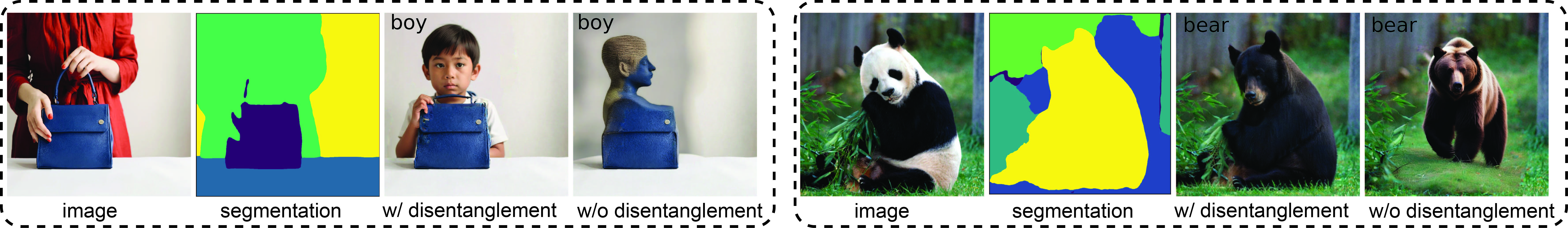}
  \caption{Qualitative comparison of textual-guided editing with and without cross-attention disentanglement}
   \vspace{-8pt}
  \label{fig::abl}
  \vspace{-10pt}
\end{figure}

We first study the influence of cross-attention disentanglement.
When the cross-attention disentanglement is not used, the learned prompt will affect the entire image and the text-guided editing will be equivalent to the legacy SDXL inpainting.
As shown in Fig. \ref{fig::abl}, when the target item and background are tightly coupled (the hand holding the bag), without disentanglement, the target prompt will take effect based on its own textual semantics and the surrounding item's semantics, therefore leading to poor editing results. 
This can be avoided by building disentangled item-prompt associations.
When the target item can be clearly separated from the background as in the panda example, introducing disentanglement can better preserve the information of the original item, making the editing more controllable.
We then study the influence of the number of tokens used to represent each item, and as demonstrated in Tab. \ref{tab::abl_tokens}, 1-5 tokens per item lead to good text-guided editing performance while too many tokens will complicate the embedding training phase and affect the results, and therefore we use 5 tokens in SDXL to generate all results.

\begin{table}[h]
\centering
\caption{Text-guided editing with different token numbers. }
\resizebox{0.6\columnwidth}{!}{
\begin{tabular}{lcccc} 
\toprule
 tokens num. & 1     & 2     & 5                & 10     \\ 
\midrule
LPIPS     & 0.183 & 0.204 & \textbf{0.179} & 0.413  \\
CLIP-T    & 38.5  & 38.2  & \textbf{42.0}  & 30.1   \\
\bottomrule
\end{tabular}
}
\vspace{-8pt}
\label{tab::abl_tokens}
\end{table}


\section{Conclusion}
In this work, we propose D-Edit as a versatile image editing framework based on diffusion models.
D-Edit segments the given image into multiple items, each of which is assigned a prompt to control its representation in the prompt space.
The image-prompt cross-attention is disentangled to a group of item-prompt interactions.
Item-prompt associations are built up by fintuning the diffusion model which learns to reconstruct the original image using the given set of item prompts.
We showcase the quality and versatility of the editing results across a diverse range of collected images through both qualitative and quantitative evaluations.

\section{Acknowledgments} This work is supported by the U.S. Department of Energy under award DE-FOA-0003264 and the Army Research Office under grant W911NF-23-1-0088.

\bibliography{aaai25}

\appendix

\onecolumn
\section{Different types of editing}

\xhdr{Text-based Item Editing.} 
Because the optimized model keeps the memory of the original vocabulary, replacing an item with another one described by a text prompt can be simply achieved by changing the corresponding item prompt to the target text. 
The learned item prompt can also be used along with other words to form a new prompt. 

\xhdr{Image-based Item Editing}
Given two images, to replace an item in the source image with another one in the reference image, all item prompts in both images should be injected into the same model. 
Then the item replacement can be realized by the corresponding prompt replacement.
Note that this setting is more general compared to classical image-based editing scenarios where the reference is an entire image instead of an item.
This is particularly useful when it is hard to find a solo focus reference.

\xhdr{Mask-based Item Editing} 
Besides replacing items by changing the prompt, we can also keep the item semantics but change the appearance of the item in the image by editing the corresponding segmentation mask.
The mask editing includes moving the mask position, resizing mask, refining the mask shape, or redrawing the mask.
Therefore, we consider four types of mask-based editing in the experiment including changing item position, size, shape, and post-editing refinement.

\xhdr{Item Removal}
Items in the image can be removed by deleting the corresponding item mask and item-prompt pair from the image.
The deleted region will be filled by nearby region masks and corresponding item-prompt pairs.

\section{Prompt Interpolation}
\begin{figure*}[h]
  \centering \includegraphics[width=0.8\linewidth]{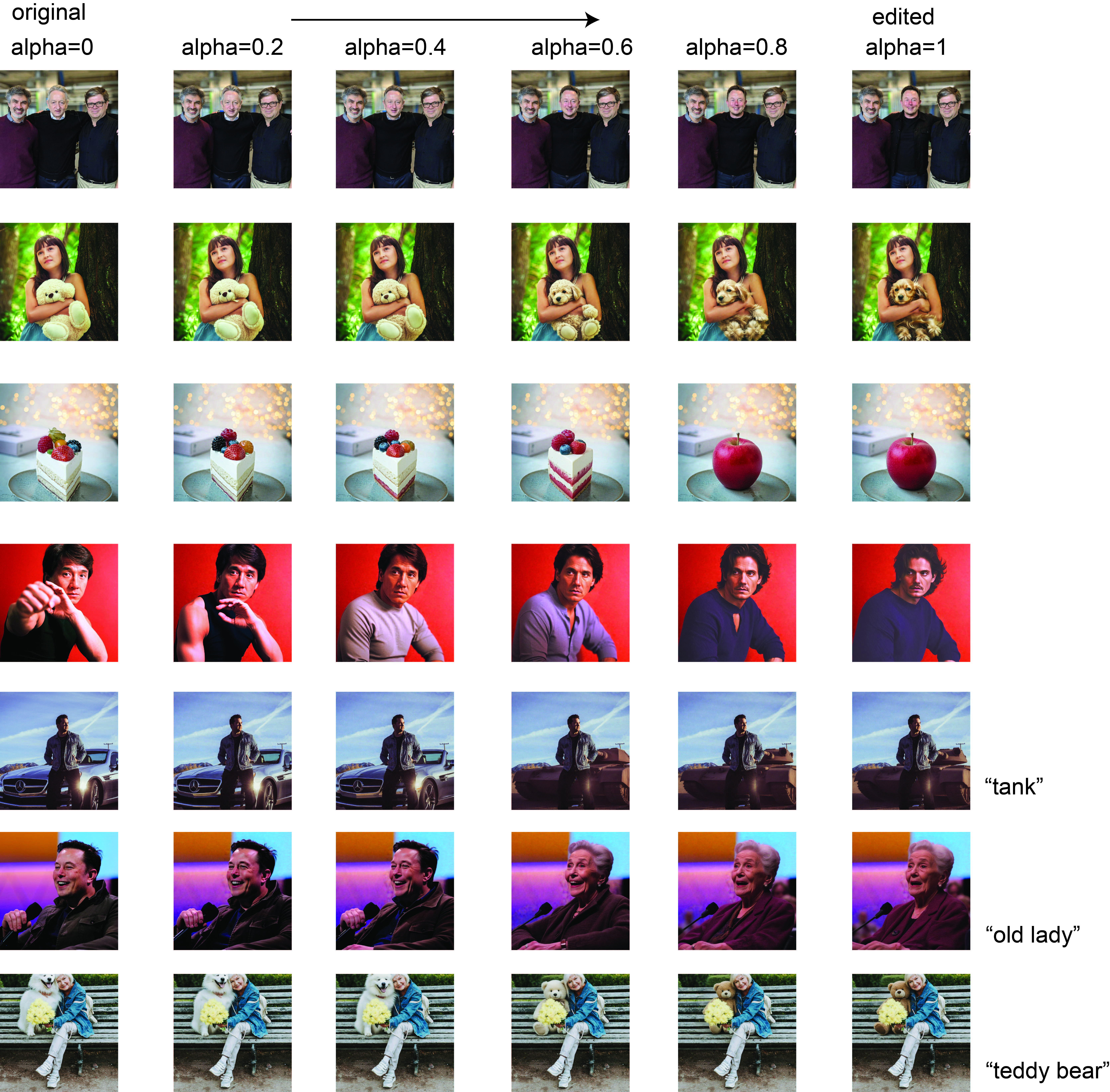}
  \caption{
  Mixing of the original and guidance images by linearly interpolating the embeddings of the two.
  }
  \label{sfig::ablate}
\end{figure*}

Instead of replacing the original item prompt with the target one (image/text guided), we can also interpolate between the two and get the transition results.
As shown in Fig. \ref{sfig::ablate}, we interpolate the embeddings after the text encoder and derive the mixed embedding using $c =\text{alpha}\cdot c_{\text{guide}} + (1-\text{alpha})\cdot c_{\text{orig}}$.

\clearpage

\section{More Text-Based Editing}
\begin{figure*}[h]
  \centering \includegraphics[width=0.8\linewidth]{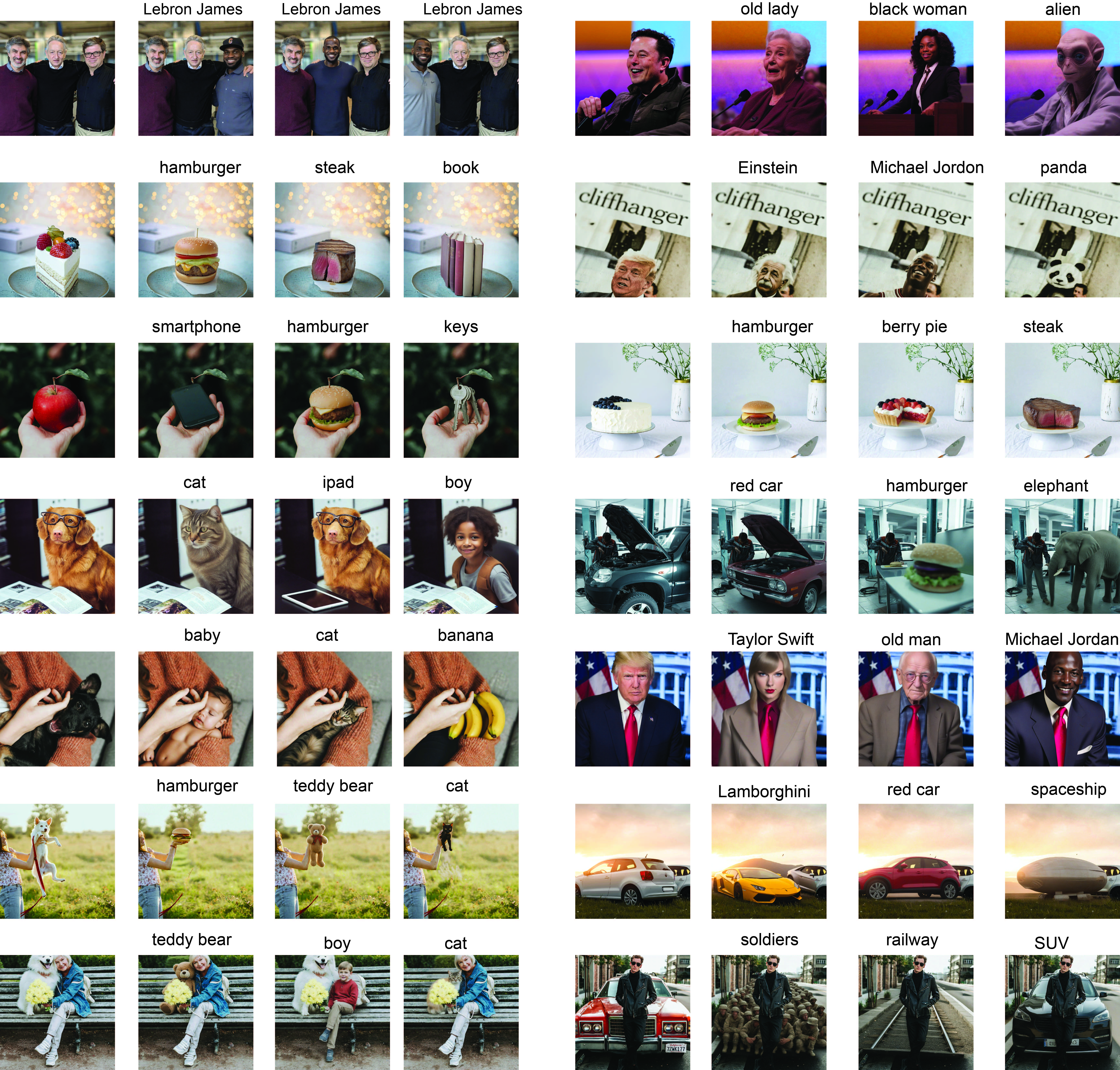}
  \caption{
  More text-editing results of replacing one item with target prompts.
  }
  \label{sfig::text}
\end{figure*}

\clearpage
\section{More Image-Based Editing}
\begin{figure*}[h]
  \centering \includegraphics[width=0.8\linewidth]{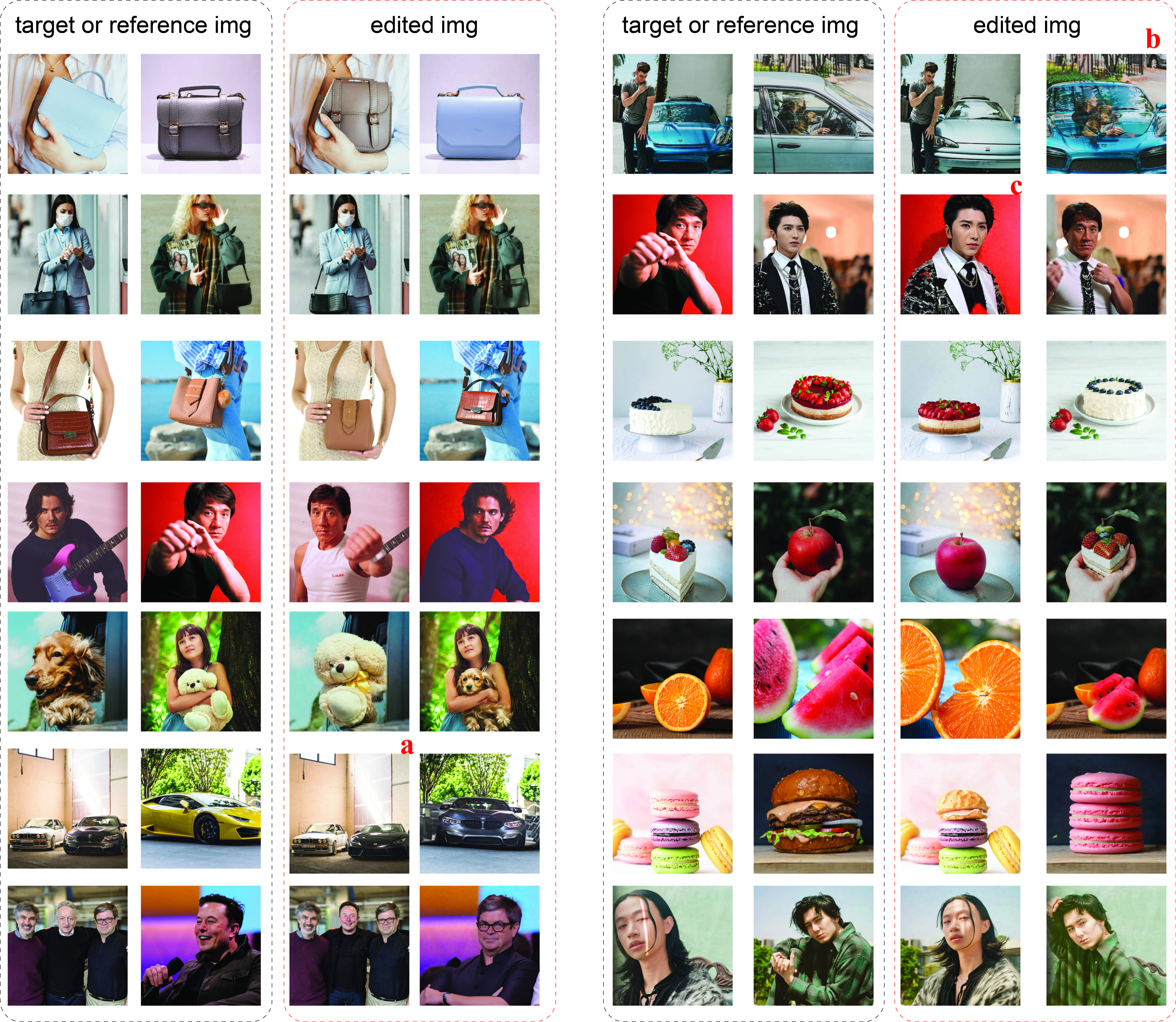}
  \caption{
  More image-guided editing results of replacing one item with the reference item.
  Failure examples are marked by red letters.
  }
  \label{sfig::img}
\end{figure*}

As marked in Fig. \ref{sfig::img}, several failure examples occur: 
(1) \textbf{Color is harder to learn than shape}: in the Lamborghini example $a$, the shape is preserved while the yellow color is lost.
(2) \textbf{Item should have clear meanings}: in the man-in-car example $b$, because the man picture through the window is blurred and DPM cannot understand, the inpainting results will be unnatural.
(3) \textbf{Out-of-domain image is hard to learn}:
In the celebrity face example $c$, if the celebrity face is not natural or highly Photoshopped, it will be hard for DPMs to reconstruct it and therefore lead to weird face results.

\end{document}